\documentclass[letterpaper, 10 pt, conference]{ieeeconf}  

\usepackage{epsfig}
\usepackage{amsmath}
\usepackage{cleveref}
\usepackage{footmisc}
\usepackage{color}
\newcommand{\Secref}[1]{Section~\ref{#1}} 
\newcommand{\Tacref}[1]{Table~\ref{#1}} 
\newcommand{\Ficref}[1]{Figure~\ref{#1}}

\IEEEoverridecommandlockouts                              

\title{\LARGE \bf
Normalization in Training U-Net for 2D Biomedical Semantic Segmentation
}

\author{Xiao-Yun Zhou$^{1}$ and Guang-Zhong Yang$^{1}$
\thanks{$^{1}$Xiao-Yun Zhou and Guang-Zhong Yang are with the Hamlyn Centre for Robotic Surgery, Imperial College London, London, UK
        {\tt\small xiaoyun.zhou14@imperial.ac.uk, g.z.yang@imperial.ac.uk}}%
}

\begin{document}

\maketitle
\thispagestyle{empty}
\pagestyle{empty}

\begin{abstract}
2D biomedical semantic segmentation is important for robotic vision in surgery. Segmentation methods based on Deep Convolutional Neural Network (DCNN) can out-perform conventional methods in terms of both accuracy and levels of automation. One common issue in training a DCNN for biomedical semantic segmentation is the internal covariate shift where the training of convolutional kernels is encumbered by the distribution change of input features, hence both the training speed and performance are decreased. Batch Normalization (BN) is the first proposed method for addressing internal covariate shift and is widely used. Instance Normalization (IN) and Layer Normalization (LN) have also been proposed. Group Normalization (GN) is proposed more recently and has not yet been applied to 2D biomedical semantic segmentation\footnote{GN was used in 3D biomedical semantic segmentation in \cite{kao2018brain}, however, no specific validations on GN were given.}. Most DCNNs for biomedical semantic segmentation adopt BN as the normalization method by default, without reviewing its performance. In this paper, four normalization methods - BN, IN, LN and GN are compared in details, specifically for 2D biomedical semantic segmentation. U-Net is adopted as the basic DCNN structure. Three datasets regarding the Right Ventricle (RV), aorta, and Left Ventricle (LV) are used for the validation. The results show that detailed subdivision of the feature map, i.e. GN with a large group number or IN, achieves higher accuracy. This accuracy improvement mainly comes from better model generalization. Codes are uploaded and maintained at Xiao-Yun Zhou's Github.
\end{abstract}

\section{Introduction}
Biomedical semantic segmentation, which labels the class/anatomy/prosthesis of each pixel/voxel in an image/volume, is important for intra-operative robotic navigation and path planning in surgery. For example, segmenting the Right Ventricle (RV) from Magnetic Resonance (MR) images intra-operatively is essential to instantiate 3D RV shapes for intra-operative navigation in robotic cardiac interventions \cite{zhou2018real}. Segmenting markers on fenestrated stent grafts is useful for instantiating 3D stent graft shapes and hence 3D robotic path planning for Fenestrated Endovascular Aortic Repair (FEVAR) \cite{zhou2018real2}.

Conventional segmentation methods for both natural and biomedical problems are usually based on features (edge, region, angle, etc.) which need an expert-designed feature extractor and classifier, while recent segmentation methods based on Deep Convolutional Neural Network (DCNN) extract and classify the features automatically with multiple non-linear modules \cite{lecun2015deep}. Fully Convolutional Network (FCN) is the first proposed DCNN which realized pixel-level classification and hence semantic segmentation by using convolutional and deconvolutional layers, as well as skip architectures \cite{long2015fully}.  Ronneberger et al. introduced FCN into 2D biomedical semantic segmentation, proposed U-Net with dense skip architectures, and achieved reasonable results on neuronal structure and cell segmentation \cite{ronneberger2015u}. A systematical review has been carried out by Litgens et al. on the application of DCNN in biomedical image analysis including segmentation, classification, detection, registration and other tasks \cite{litjens2017survey}. DCNNs for biomedical semantic segmentation could be divided into 3D \cite{li20183d} and 2D \cite{bernard2018deep} based on the dimension of convolution. In this paper, 2D DCNNs are discussed.

Most of previous research on DCNNs for biomedical semantic segmentation focused on architecture design, loss function, and network cascade for specific tasks. For example, atriaNet composed of multi-scaled and dual-pathed convolutional architectures was proposed for left atrial segmentation from late gadolinium enhanced MR Imaging \cite{xiong2018fully}. A hierarchical DCNN was designed with a two-stage FCN and dice-sensitivity-like loss function to segment breast tumours from dynamic contrast-enhanced MR imaging \cite{zhang2018hierarchical}. The thrombus was segmented from Computed Tomography (CT) images with detectnet, FCN and holistically-nested edge detection \cite{lopez2018fully}. Equally-weighted Focal U-Net combined with focal loss and U-Net was proposed to segment the small metal markers from fluoroscopic images of fenestrated stent grafts \cite{zhou2018towards}.

 The main step in DCNN is to apply convolutional kernels with trainable parameters on feature maps to extract new features iteratively. During the training of a DCNN, the input of a layer depends on all the parameters/values in its previous layers/feature maps. Small changes in shallow input feature maps or image batches accumulate and amplify along the depth of network, causing deep layers to be trained to fit these distribution changes rather than the real and useful content. This phenomenon is called internal covariate shift \cite{ioffe2015batch}. Batch Normalization (BN) \cite{ioffe2015batch} is the first proposed solution for internal covariate shift which normalized the feature map along the channel direction and remained the representation capacity of DCNN by re-scaling and re-translating the normalized feature map. It is the most widely used DCNN normalization method for both biomedical and natural semantic segmentation.
 
 Later Instance Normalization (IN) was proposed to normalize the feature map along both the batch and channel dimension for image stylization \cite{ulyanov1607instance}. Layer Normalization (LN) was proposed to normalize the feature map along the batch dimension for recurrent neural network \cite{ba2016layer}. Group Normalization (GN) was proposed to normalize the feature map along the batch and divided channel dimension for image classification and instance segmentation\footnote{"Instance" segmentation which labels both the class and the instance of a pixel is different from semantic segmentation.} \cite{wu2018group}. Weight normalization was proposed to normalize the trainable weights with validations on supervised image recognition, generative modelling and deep reinforcement learning, its accuracy improvement is similar to BN in terms of image classification \cite{salimans2016weight}. Cosine normalization was proposed to use cosine similarity instead of dot product in DCNN to decrease the internal covariate shift with limited validations shown in \cite{luo2017cosine}. It is complex to implement cosine normalization, as it needs to rewrite convolution and deconvolution functions.

\begin{figure}[thpb]
\centering
\framebox{\includegraphics[scale=0.27]{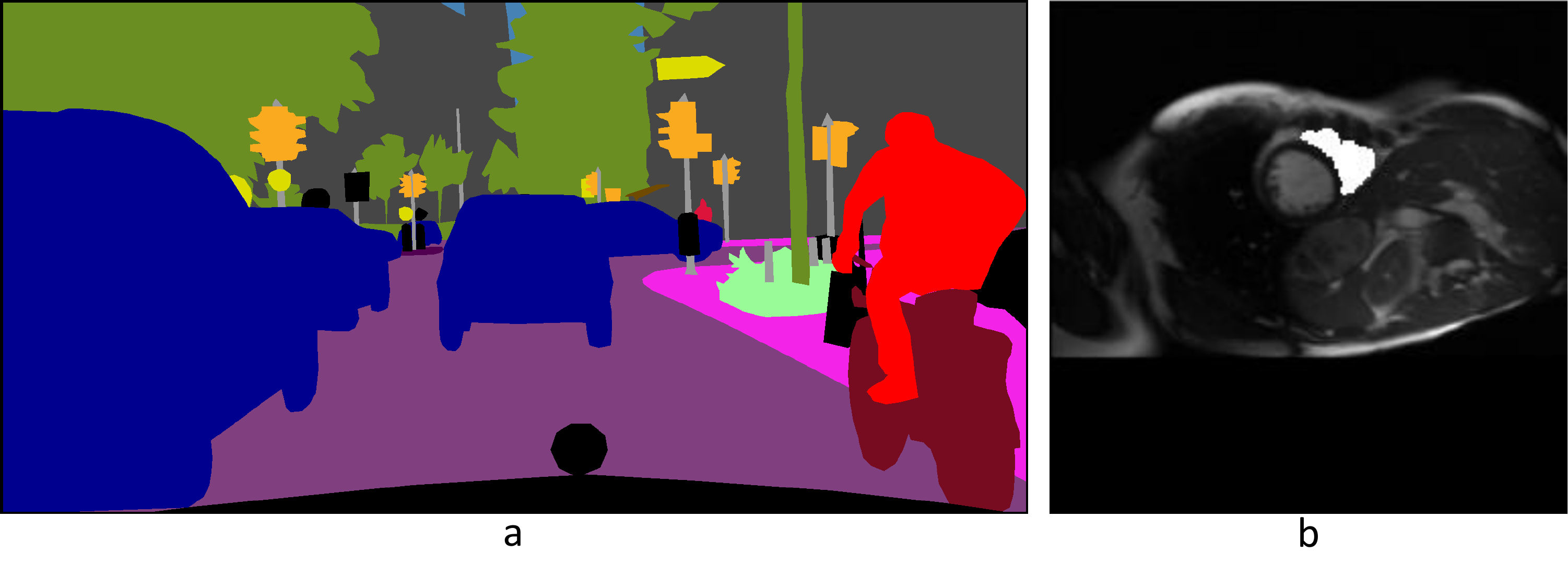}}
\caption{(a) semantic segmentation of cars, people, trees, etc. from a natural image \cite{cordts2016cityscapes}, (b) semantic segmentation of RV from a MR image.}
\label{fig:intro}
\end{figure}

These normalization methods are proposed for different tasks and there are thus far no comparisons regarding their performance. The comparisons in \cite{wu2018group} between BN, IN, LN and GN/BN and GN are for image classification/instance segmentation. In natural semantic segmentation (\Ficref{fig:intro}a), tasks are universal, indicating shareable parameters. Fine-tuning or extracting features from pre-trained feature maps are popular, preventing from exploring different normalization methods. A BN is used by default without comparing its performance with other normalization methods. In biomedical semantic segmentation (\Ficref{fig:intro}b), the target is a specific anatomy or prosthesis. A network trained from scratch is common, memory-efficient and accurate, allowing exploring different normalization methods.

In this paper, the most widely applied four normalization methods - BN, IN, LN, and GN are reviewed and compared specifically for biomedical semantic segmentation. U-Net is selected as the network architecture due to its wide application. The U-Net details, four normalization methods, data collection for the RV, aorta and Left Ventricle (LV), and the implementation details are introduced in \Secref{sec:method}. Detailed experiments and comparisons are provided in \Secref{sec:result}. It is proved that detailed subdivision of the feature map, i.e. GN with a large group number or IN, out-performed other normalization methods in terms of accuracy, despite the fact that BN is currently widely used. No obvious improvements regarding the convergence speed and lowest loss are observed. Discussion and conclusion are in \Secref{sec:discuss} and \Secref{sec:conclusion} respectively.

\section{METHODOLOGY}
\label{sec:method}
Systematic details about DCNN can be found in \cite{sze2017efficient}, while this paper focuses on explaining the concepts of data propagation, network architecture and loss function in \Secref{sec:network}. The algorithms of BN, IN, LN and GN are explained in \Secref{sec:BN}, \Secref{sec:IN}, \Secref{sec:LN} and \Secref{sec:GN} respectively. The data collection and implementation details are given in \Secref{sec:data}.

\subsection{Network Details}
\label{sec:network}
With an input feature map $F_{\rm N\times \rm H\times \rm W \times \rm C}$ (the first feature map is the image batch), $\rm N$ is the batch size, $\rm H$ the height, $\rm W$ the width, $\rm C$ the channel, a trainable convolutional kernel $T_{\rm C \times \rm K \times \rm K}$ moves along the height and width of $F_{\rm N\times \rm H\times \rm W \times \rm C}$, indicating an output feature map:

\begin{equation}
\hat{F}_{\rm N\times \rm H'\times \rm W' \times \rm 1} = F_{\rm N\times \rm H\times \rm W \times \rm C} \cdot  T_{\rm C \times \rm K \times \rm K}
\end{equation}
Where $\rm K$ is the convolutional kernel size, $\rm H'=H//S$, $\rm W'=W//S$, where $//$ is exact division and  $\rm S$ is the convolutional stride. When $\rm S>1$, the feature spatial dimension decreases. When $\rm 0<S<1$, the feature spatial dimension increases. For extracting richer features, multiple $T_{\rm C \times \rm K \times \rm K}$ are trained, resulting in $\hat{F}_{\rm N\times \rm H'\times \rm W' \times \rm C'}$.

\begin{figure*}[thpb]
\centering
\framebox{\includegraphics[scale=0.8]{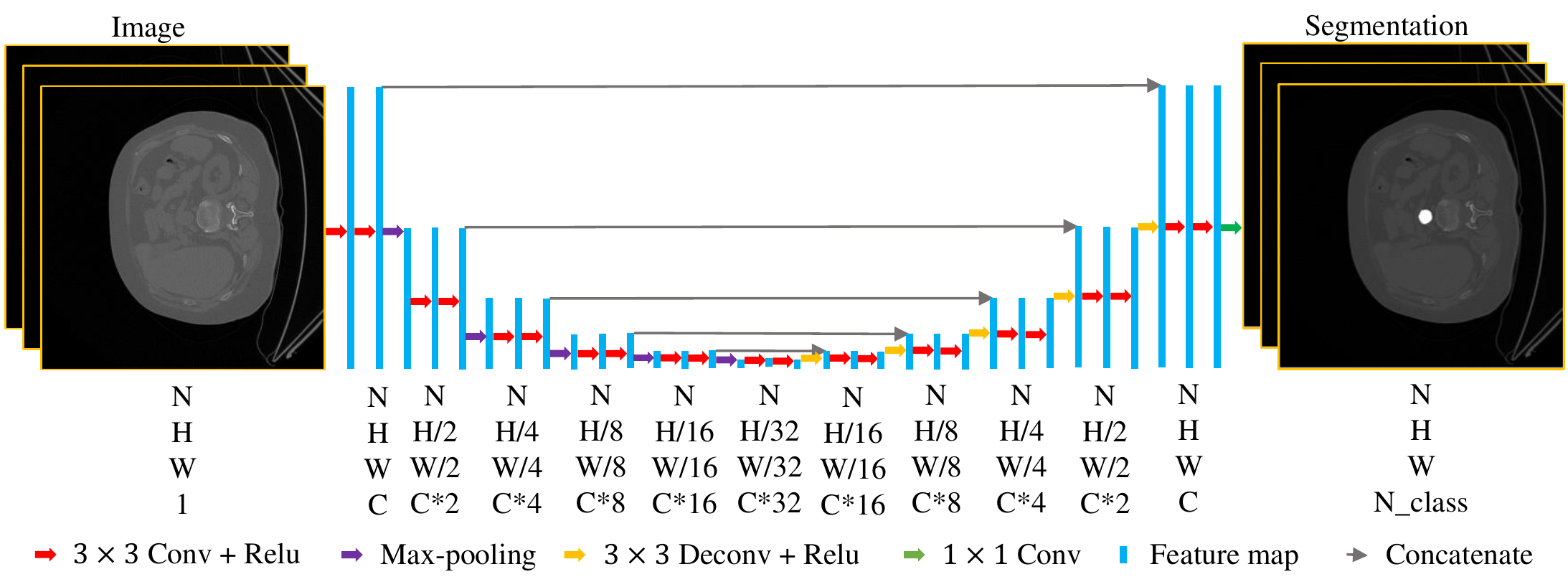}}
\caption{The structure of U-Net used in this paper, Conv - convolution, Deconv - deconvolution.}
\label{fig:structure}
\end{figure*}

U-Net, which is a widely applied DCNN structure for biomedical semantic segmentation is used as the network architecture in this paper, its architecture is shown in \Ficref{fig:structure}. It gradually increases the receptive field (the pixels it sees) with max-pooling layers, resulting in decreased spatial dimensions. Then U-Net recovers and increases the spatial dimension with deconvolutional layers.

An increased receptive field is useful for extracting the semantic information. However, the consequent decreased spatial dimension is disadvantageous for spatial information. Skip connections are used to concatenate the feature maps from shallow layers and deep layers to combine the semantic and spatial information.

All the convolutional and deconvolutional layers are followed with a Relu activation, except the $1\times 1$ convolutional layer at the last which predicts the class probability. Softmax is used to transform the final feature map into probabilities and cross-entropy is used as the loss function:

\begin{equation}
  loss(p,y) =
  \begin{cases}
    -log(p) & \text{if $y = 1.0$} \\
    -log(1.0-p) & \text{if $y = 0.0$} \\
  \end{cases}
\end{equation}

Where $y$ is the ground truth, $p$ is the prediction probability. Stochastic Gradient Descent (SGD) is adopted to train the $T_{\rm C \times \rm K \times \rm K \times \rm C'}$ to obtain a minimum loss. When the distribution of $F_{\rm N\times \rm H\times \rm W \times \rm C}$ changes, $T_{\rm C \times \rm K \times \rm K \times \rm C'}$ is influenced and trained to fit this distribution change, resulting in interval covariate shift which decreases both the training speed and accuracy. 

\subsection{Batch Normalization (BN)}
\label{sec:BN}
BN is the first proposed algorithm for solving the interval covariate shift. It normalizes $F_{\rm N\times \rm H\times \rm W \times \rm C}$ with the mean of 0.0 and the variance of 1.0 while maintains the representation capability of a DCNN with two more trainable parameters - $\gamma, \beta$.

In BN \cite{ioffe2015batch}, the mean and variance are calculated along the channel:
\begin{equation}
\mu_c = \frac{1}{\rm N \times \rm H \times \rm W} \sum\limits_{n=1}^{\rm N} \sum\limits_{h=1}^{\rm H} \sum\limits_{w=1}^{\rm W} f_{n,h,w}
\end{equation}

\begin{equation}
\delta_c^2 = \frac{1}{\rm N \times \rm H \times \rm W} \sum\limits_{n=1}^{\rm N} \sum\limits_{h=1}^{\rm H} \sum\limits_{w=1}^{\rm W} (f_{n,h,w}-\mu_c)^2
\end{equation}

The feature map is normalized by:
\begin{equation}
\hat{f}_{n,h,w} = \frac{f_{n,h,w}-\mu_c}{\sqrt{\delta_c^2+\epsilon}}
\end{equation}
where $\epsilon$ is a small value used to increase the division stability. After this normalization, $\hat{f}_{n,h,w}$ is always with the mean of 0.0 and the variance of 1.0, which limits the DCNN representation capacity. Additional trainable parameters $\gamma_c$ and $\beta_c$ are added to each channel to recover the representation power:
\begin{equation}
f'_{n,h,w} = \gamma_c \hat{f}_{n,h,w}+\beta_c
\end{equation}

BN is applied after the convolution and before the activation. There are two ways of applying BN during the inference: 1) use the moving average mean and variance in the training stage to normalize the test feature map, recommended in \cite{ioffe2015batch}; 2) use the mean and variance in the test stage to normalize the test feature map, recommended in \cite{ulyanov1607instance}. In this paper, both ways are explored and the optimal one is used for the comparison with other normalization methods.

\subsection{Instance Normalization (IN)}
\label{sec:IN}
In IN \cite{ulyanov1607instance}, the mean and variance are calculated along the channel and batch:
\begin{equation}
\mu_{n,c} = \frac{1}{\rm H \times \rm W} \sum\limits_{h=1}^{\rm H} \sum\limits_{w=1}^{\rm W} f_{h,w}
\end{equation}

\begin{equation}
\delta_{n,c}^2 = \frac{1}{\rm H \times \rm W} \sum\limits_{h=1}^{\rm H} \sum\limits_{w=1}^{\rm W} (f_{h,w}-\mu_{n,c})^2
\end{equation}
The feature map is normalized by:
\begin{equation}
\hat{f}_{h,w} = \frac{f_{h,w}-\mu_{n,c}}{\sqrt{\delta_{n,c}^2+\epsilon}}
\end{equation}

\subsection{Layer Normalization (LN)}
\label{sec:LN}
In LN \cite{ba2016layer}, the mean and variance are calculated along the batch:
\begin{equation}
\mu_n = \frac{1}{\rm H \times \rm W \times \rm C } \sum\limits_{h=1}^{\rm H} \sum\limits_{w=1}^{\rm W} \sum\limits_{c=1}^{\rm C} f_{h,w,c}
\end{equation}

\begin{equation}
\delta_n^2 = \frac{1}{\rm H \times \rm W \times \rm C} \sum\limits_{h=1}^{\rm H} \sum\limits_{w=1}^{\rm W} \sum\limits_{c=1}^{\rm C} (f_{h,w,c}-\mu_n)^2
\end{equation}
The feature map is normalized by:
\begin{equation}
\hat{f}_{h,w,c} = \frac{f_{h,w,c}-\mu_n}{\sqrt{\delta_n^2+\epsilon}}
\end{equation}

\subsection{Group Normalization (GN)}
\label{sec:GN}
In GN \cite{wu2018group}, the mean and variance are calculated along the batch and divided channel. The difference between GN and IN/LN is that a group of channels $\rm M=C//G$ are grouped together for the normalization, $\rm G$ is the group number, $\rm M$ is the channel number per group:
\begin{equation}
\mu_{n,g} = \frac{1}{\rm H \times \rm W \times \rm M} \sum\limits_{h=1}^{\rm H} \sum\limits_{w=1}^{\rm W} \sum\limits_{m=(g-1)\cdot M+1}^{\rm g\cdot M} f_{h,w,m}
\end{equation}

\begin{equation}
\delta_{n,g}^2 = \frac{1}{\rm H \times \rm W \times \rm M} \sum\limits_{h=1}^{\rm H} \sum\limits_{w=1}^{\rm W} \sum\limits_{m=(g-1)\cdot M+1}^{\rm g\cdot M} (f_{h,w,m}-\mu_{n,g})^2
\end{equation}

The feature map is normalized by:
\begin{equation}
\hat{f}_{h,w,m} = \frac{f_{h,w,m}-\mu_{n,g}}{\sqrt{\delta_{n,g}^2+\epsilon}}
\end{equation}

In this paper, GN with different group numbers is seen as different normalization methods and are compared.

In IN, LN, and GN, additional parameters are also used for recovering the DCNN representation ability. Multiple ways of adding $\gamma$ and $\beta$ may exist. In this paper, we follow \cite{wu2018group} and add parameters for each feature channel, which results in $\rm 2C$ parameters for each feature map. 

\subsection{Data Collection and Implementation Details}
\label{sec:data}
Three datasets: RV scanned with MR, with $256\times 256$ image size, aorta scanned with CT, with $512\times 512$ image size, and LV scanned with MR, with $256\times 256$ image size are used for the validation.

37 RV scans \cite{lekadir2007outlier} were acquired from a 1.5T MR scanner (Sonata, Siemens, Erlangen, Germany), from both the asymptomatic and Hypertrophic Cardiomyopathy (HCM) subjects, from the atrioventricular ring to the apex, with a $10mm$ slice gap, a $1.5-2mm$ pixel spacing, and $19-25$ time frames for the cardiac cycle. 6082 images were collected in total. All images were labelled by one expert with Analyze (AnalyzeDirect, Inc, Overland Park, KS, USA) and were augmented by rotation from $-30^\circ$ to $30^\circ$ with $10^\circ$ as the interval. The 37 subjects were split randomly into three groups for cross validation, with 12, 12, and 13 subjects for each group respectively. 

20 aortic CT scans were acquired from the VISCERAL data set \cite{jimenez2016cloud}. 4631 images were collected in total and were augmented by rotation from $-40^\circ$ to $40^\circ$ with $10^\circ$ as the interval. The 20 subjects were split randomly into three groups for cross validation, with 7, 7, and 6 subjects for each group respectively.

45 LV MR scans were acquired from the SunnyBrook data set \cite{radau2009evaluation}. 805 images were collected in total and were augmented by rotation from $-60^\circ$ to $60^\circ$ with $2^\circ$ as the interval. The 45 subjects were split randomly into three groups for cross validation, with 15 subjects for each group.

All image intensities were re-scaled to a maximum value of 1.0 and a minimum value of 0.0. In the cross validation, one group was used as the testing data while the other two groups were used for the training data. No evaluation images were split.

The kernel size of convolutional and deconvolutional layers is 3, except the last convolutional layer whose kernel size is 1. The pool size for max-pooling is 2. The stride of deconvolutional layers is 2. The root number of feature map - C is 16. The moment is 0.9. Step-wise learning rate schedule was used, as it allows careful and manual adjustment of the learning rate. Training with two epochs usually achieved the lowest loss and was used in our experiments. Three to four initial learning rates were tested for each experiment and the one with the best performance is reported in this paper. Several step-wise methods were explored, i.e. dividing the learning rate by 5 or 10 every half or one epoch. Dividing the learning rate by 5 at the second epoch showed optimal performance and was used. This is also consistent with the learning rate schedule in \cite{smith2017don}.

The largest explored batch size for the RV, aorta, and LV are 32, 16, and 32 respectively in this paper. This is determined by the GPU memory. As the RV and LV are with smaller image size and consume less GPU memories, the largest batch size the GPU can hold is larger.

The DCNN framework is programmed with Tensorflow Estimator Application Programming Interface (API) tersely, with about 200 lines of codes. The convolutional, deconvolutional, and max-pooling layers are programmed with tf.layers. The programming of BN is from tf.layers while that of IN, LN and GN are from tf.contrib.layers. The data is fed into Tensorflow with tfrecords and tf.data API. The images were shuffled when generating the tfrecords files and then were shuffled again with a shuffle size of 500 during the feeding, which guaranteed random input images.

Dice Similarity Coefficient (DSC) was calculated as the evaluation metric:
\begin{equation}
DSC = 2\cdot \frac{\rm Y \cap \rm P}{\rm Y + \rm P}
\end{equation}
where $\rm Y$ is the ground truth and $\rm P$ is the prediction. As only two classes exist, the trend of background DSC is the same as that of the foreground DSC, hence only the foreground DSC is shown in this paper. 

\begin{table}
\centering
\caption{Mean$\pm$std DSCs of segmenting the RV, aorta, and LV with BN; TestI and TrainI are used during inference; "-" means an optimal LR could not be found for that case.}
\begin{tabular}{|c|c|cc|cc|}
\hline
BS  & Test & \multicolumn{2}{c|}{Mean$\pm$std DSCs} & \multicolumn{2}{c|}{Optimal LR}\\
   &   & TestI & TrainI & TestI & TrainI\\
\hline
S  & RV-1    & \textcolor{blue}{\textbf{0.7133$\pm$0.2693}} & 0.6895$\pm$0.2760 & 0.5 & 0.05\\
       & RV-2    & \textcolor{blue}{\textbf{0.7139$\pm$0.2859}} & 0.6579$\pm$0.3306 & 1.0 & 0.1\\
       & RV-3    & \textcolor{blue}{\textbf{0.6745$\pm$0.3029}} & 0.6070$\pm$0.3390 & 1.0 & 0.05\\
       &aorta-1  & \textcolor{blue}{\textbf{0.8368$\pm$0.1405}} & 0.8249$\pm$0.1773 & 0.5 & 0.5\\
       &aorta-2  & 0.7689$\pm$0.2178 & \textcolor{blue}{\textbf{0.7832$\pm$0.2072}} & 0.5 & 0.1\\
       &aorta-3  & \textcolor{blue}{\textbf{0.8060$\pm$0.2294}} & 0.7707$\pm$0.2707 & 1.5 & 0.1\\
       &LV-1     & \textcolor{blue}{\textbf{0.9240$\pm$0.0808}} & 0.9020$\pm$0.1110 & 0.5 & 0.1\\
       &LV-2     & \textcolor{blue}{\textbf{0.8864$\pm$0.1391}} & 0.8686$\pm$0.1999 & 1.0 & 0.1\\
       &LV-3     & \textcolor{blue}{\textbf{0.8479$\pm$0.1643}} & 0.8063$\pm$0.2300 & 1.0 & 0.1\\
M & RV-1    & \textcolor{blue}{\textbf{0.7025$\pm$0.2796}} & 0.6533$\pm$0.2930 & 1.0 & 0.1\\
       & RV-2    & \textcolor{blue}{\textbf{0.6833$\pm$0.3091}} & 0.6131$\pm$0.3316 & 1.0 & 0.1\\
       & RV-3    & \textcolor{blue}{\textbf{0.6415$\pm$0.3275}} & 0.5529$\pm$0.3678 & 1.0 & 0.05\\
       & aorta-1 & 0.7804$\pm$0.2061 & \textcolor{blue}{\textbf{0.8036$\pm$0.1714}} & 1.5 & 0.1\\
       & aorta-2 & 0.7276$\pm$0.2525 & \textcolor{blue}{\textbf{0.7726$\pm$0.2009}} & 1.0 & 0.5\\
       & aorta-3 & 0.7408$\pm$0.2798 & \textcolor{blue}{\textbf{0.7787$\pm$0.2453}} & 1.0 & 0.5\\
       & LV-1    & \textcolor{blue}{\textbf{0.9054$\pm$0.0864}} & 0.8384$\pm$0.2111 & 0.5 & 0.1\\
       & LV-2    & 0.8431$\pm$0.1769 & \textcolor{blue}{\textbf{0.8567$\pm$0.1815}} & 0.5 & 0.5\\
       & LV-3    & \textcolor{blue}{\textbf{0.7899$\pm$0.2186}} & 0.7085$\pm$0.2701 & 1.0 & 0.05\\
L  & RV-1    & \textcolor{blue}{\textbf{0.6794$\pm$0.2847}} & 0.6556$\pm$0.2873 & 1.0 & 0.5\\
       & RV-2    & \textcolor{blue}{\textbf{0.6670$\pm$0.3066}} & 0.6283$\pm$0.3108 & 1.0 & 0.5\\
       & RV-3    & \textcolor{blue}{\textbf{0.6380$\pm$0.3267}} & 0.5838$\pm$0.3377 & 0.5 & 0.1\\
       & aorta-1 & 0.7668$\pm$0.2070 & \textcolor{blue}{\textbf{0.7782$\pm$0.1842}} & 1.0 & 0.1\\
       & aorta-2 & 0.7200$\pm$0.2423 & \textcolor{blue}{\textbf{0.7458$\pm$0.2243}} & 1.5 & 0.1\\
       & aorta-3 & 0.6200$\pm$0.3557 & \textcolor{blue}{\textbf{0.7449$\pm$0.2594}} & 1.5 & 0.1\\
       & LV-1    & \textcolor{blue}{\textbf{0.8868$\pm$0.1432}} & - & 0.5 & -\\
       & LV-2    & \textcolor{blue}{\textbf{0.7892$\pm$0.2325}} & 0.7085$\pm$0.1887 & 1.0 & 0.1\\
       & LV-3    & \textcolor{blue}{\textbf{0.7677$\pm$0.2151}} & 0.7076$\pm$0.2919 & 0.5 & 0.5\\
\hline
\end{tabular}
\label{tab:inference} 
\end{table}

\begin{table*}
\centering
\caption{Mean$\pm$std DSCs of the RV segmentation with different normalization methods (highest DSC in bold and blue).}
\begin{tabular}{|c|c|ccc|ccc|ccc|}
\hline
Experiment &Normalization &\multicolumn{3}{c|}{Mean$\pm$std DSCs} &\multicolumn{3}{c|}{Optimal LR} &\multicolumn{3}{c|}{Time (Seconds/20 Iterations)}\\
           &               &S-BS  &M-BS  &L-BS                    &S-BS &M-BS &L-BS                           &S-BS &M-BS &L-BS   \\
\hline
RV-1 &None &0.6944$\pm$0.2428 &0.6654$\pm$0.2838 &0.6306$\pm$0.2647 &0.5 &0.1 &1.0 &0.5 &2.6 &4.8\\
&BN &0.7133$\pm$0.2693 &\textcolor{blue}{\textbf{0.7025$\pm$0.2796}} &\textcolor{blue}{\textbf{0.6794$\pm$0.2847}} &0.5 &1.0 &1.0 &0.88 &3.6 &6.6\\
&GN4 &0.7023$\pm$0.3078 &0.6887$\pm$0.2734 &0.6791$\pm$0.2879 &0.5 &0.5 &1.0 &1.1 &5.3 &9.4\\
&GN8 &0.6952$\pm$0.2932 &0.6744$\pm$0.3033 &0.6616$\pm$0.3098 &1.5 &0.5 &1.5 &1.1 &5.2 &9.3\\
&GN16 &0.6989$\pm$0.2964 &0.6755$\pm$0.2966 &0.6732$\pm$0.2991 &1.0 &0.5 &1.0  &1.1 &5.3 &9.2\\
&IN &\textcolor{blue}{\textbf{0.7346$\pm$0.2352}} &0.6856$\pm$0.2927 &0.6662$\pm$0.3121 &1.5 &1.5 &1.0 &1.0 &4.0 &7.3\\
&LN &0.6906$\pm$0.2954 &0.6928$\pm$0.2701 &0.6606$\pm$0.2876 &0.1 &0.5 &0.5 &1.0 &4.1 &7.5\\
\hline     
RV-2 &None &0.6452$\pm$0.3297 &0.6150$\pm$0.3130 &0.5951$\pm$0.3339 &0.1 &0.5 &0.1 &0.5 &2.6 &4.8\\
&BN &0.7139$\pm$0.2859 &0.6833$\pm$0.3091 &0.6670$\pm$0.3066 &1.0 &1.0 &1.0 &0.88 &3.6 &6.6\\
&GN4 &0.6795$\pm$0.3088 &0.6238$\pm$0.3424 &0.6439$\pm$0.3136 &0.1 &1.0 &0.1 &1.1 &5.3 &9.4\\
&GN8 &0.7155$\pm$0.2752 &0.6258$\pm$0.3503 &0.6386$\pm$0.3253 &1.0 &0.1 &1.0 &1.1 &5.2 &9.3\\
&GN16 &\textcolor{blue}{\textbf{0.7291$\pm$0.2720}} &\textcolor{blue}{\textbf{0.6835$\pm$0.3091}} &\textcolor{blue}{\textbf{0.6785$\pm$0.3027}} &0.5 &0.5 &1.5 &1.1 &5.3 &9.2\\
     & IN            & 0.7022$\pm$0.3002 & 0.6565$\pm$0.3303 & 0.6382$\pm$0.3448 & 0.1 & 1.0 & 1.0 & 1.0 & 4.0 & 7.3\\
     & LN            & 0.6789$\pm$0.3123 & 0.6376$\pm$0.3175 & 0.6418$\pm$0.3151 & 0.5 & 1.0 & 0.1 & 1.0 & 4.1 & 7.5\\
\hline     
RV-3 & None          & 0.6117$\pm$0.3455 & 0.5715$\pm$0.3490 & 0.5629$\pm$0.3358&0.05 & 0.1 & 0.05 & 0.5 & 2.6 & 4.8\\
     & BN            & 0.6745$\pm$0.3029 & 0.6415$\pm$0.3275 & 0.6380$\pm$0.3267 & 1.0 & 1.0 & 0.5 & 0.88 & 3.6 & 6.6\\
     & GN4           & 0.6548$\pm$0.2963 & 0.5931$\pm$0.3520 & 0.5850$\pm$0.3474 & 0.5 & 0.5 & 0.5 & 1.1 & 5.3 & 9.4\\
     & GN8           & 0.6366$\pm$0.3354 & 0.6092$\pm$0.3459 & 0.5867$\pm$0.3495 & 1.0 & 1.0 & 0.5 & 1.1 & 5.2 & 9.3\\
     & GN16 &0.6735$\pm$0.3080 & \textcolor{blue}{\textbf{0.7153$\pm$0.2683}} & \textcolor{blue}{\textbf{0.6532$\pm$0.3161}}&1.0 & 1.5 & 0.5 & 1.1 & 5.3 & 9.2\\
     & IN &\textcolor{blue}{\textbf{0.7145$\pm$0.2732}} & 0.6317$\pm$0.3282 & 0.6158$\pm$0.3391&1.0 & 0.5 & 1.0 & 1.0 & 4.0 & 7.3\\
     & LN            & 0.6118$\pm$0.3366 & 0.6292$\pm$0.3348 & 0.6025$\pm$0.3316 & 0.1 & 1.0 & 0.05& 1.0 & 4.1 & 7.5\\
\hline
\end{tabular}
\label{tab:RV} 
\end{table*}

\begin{table*}
\centering
\caption{Mean$\pm$std DSCs of the aorta segmentation with different normalization methods (highest DSC in bold and blue).}
\begin{tabular}{|c|c|ccc|ccc|ccc|}
\hline
Experiment & Normalization & \multicolumn{3}{c|}{Mean$\pm$std DSCs} & \multicolumn{3}{c|}{Optimal LR} & \multicolumn{3}{c|}{Time (Seconds/20 Iterations)}\\
           &               & S-BS  &M-BS  &L-BS                     & S-BS & M-BS & L-BS                         & S-BS & M-BS & L-BS   \\
\hline
Aorta-1 & None          & 0.8165$\pm$0.1843 & 0.7965$\pm$0.1689 & 0.7932$\pm$0.2030 & 0.05 & 0.05 & 0.5 & 1.0 & 7.9 & 15.0\\
        & BN            & 0.8368$\pm$0.1405 & 0.8036$\pm$0.1714 & 0.7782$\pm$0.1842 & 0.5 & 1.5 & 1.0   & 1.5 & 6.5 & 12.5\\
        & GN4           & 0.8310$\pm$0.1556 & 0.7928$\pm$0.1783 & 0.7615$\pm$0.2122 & 1.5 & 0.1 & 1.0   & 2.2 & 10.5 & 18.5\\
        & GN8           & 0.8314$\pm$0.1620 & 0.8223$\pm$0.1745 & 0.8065$\pm$0.1496 & 1.0 & 1.0 & 1.0   & 2.0 & 10.7 & 19.0\\
        & GN16 & \textcolor{blue}{\textbf{0.8412$\pm$0.1483}} & 0.8207$\pm$0.1613 & 0.8155$\pm$0.1431 & 0.5 & 1.5 & 1.0  & 1.7 & 11.5 & 19.2\\
        & IN & 0.8320$\pm$0.1518 & \textcolor{blue}{\textbf{0.8273$\pm$0.1313}} & \textcolor{blue}{\textbf{0.8174$\pm$0.1478}}&1.5 & 1.0 & 1.0 & 1.7 & 7.1 & 13.7\\
        & LN            & 0.8193$\pm$0.1913 & 0.7292$\pm$0.2391 & 0.7038$\pm$0.2930 & 1.0 & 1.0 & 0.5 & 1.6 & 7.4 & 14.3\\
\hline      
Aorta-2 & None & 0.7938$\pm$0.2081 & 0.7692$\pm$0.2175 & 0.7532$\pm$0.2480&0.05 & 0.1 & 0.5     & 1.0 & 7.9 & 15.0\\
        & BN            & 0.7832$\pm$0.2072 & 0.7726$\pm$0.2009 & 0.7458$\pm$0.2243 & 0.5 & 1.0 & 1.5   & 1.5 & 6.5 & 12.5\\
        & GN4           & 0.7863$\pm$0.2090 & 0.7681$\pm$0.2201 & 0.6923$\pm$0.2935 & 1.0 & 1.0 & 0.1   & 2.2 & 10.5 & 18.5\\
        & GN8 & \textcolor{blue}{\textbf{0.8099$\pm$0.1724}} & 0.7588$\pm$0.2425 & \textcolor{blue}{\textbf{0.7623$\pm$0.2275}} & 0.5 & 0.5 & 0.5& 2.0 & 10.7 & 19.0\\
        & GN16 & 0.7916$\pm$0.1974 & \textcolor{blue}{\textbf{0.7891$\pm$0.1934}} & 0.7084$\pm$0.2587&0.5 & 1.5 & 0.5 & 1.7 & 11.5 & 19.2\\
        & IN            & 0.7734$\pm$0.2166 & 0.7529$\pm$0.2278 & 0.7249$\pm$0.2663 & 1.0 & 1.0 & 1.5   & 1.7 & 7.1 & 13.7\\
        & LN            & 0.7793$\pm$0.2213 & 0.7270$\pm$0.2623 & 0.7053$\pm$0.2339 & 1.0 & 1.0 & 1.5   & 1.6 & 7.4 & 14.3\\
\hline     
Aorta-3 & None          & 0.7718$\pm$0.2712 & 0.7611$\pm$0.2821 & 0.7281$\pm$0.2798 & 0.05 & 0.5 & 0.5  & 1.0 & 7.9 & 15.0\\
        & BN            & 0.8060$\pm$0.2294 & 0.7787$\pm$0.2453 & 0.7449$\pm$0.2594 & 1.5 & 1.0 & 1.5   & 1.5 & 6.5 & 12.5\\
        & GN4           & 0.7917$\pm$0.2654 & 0.7423$\pm$0.2810 & 0.7217$\pm$0.3010 & 0.5 & 0.5 & 0.1   & 2.2 & 10.5 & 18.5\\
        & GN8 & 0.8059$\pm$0.2490 & 0.7715$\pm$0.2632 & \textcolor{blue}{\textbf{0.7774$\pm$0.2420}} & 0.5 & 1.0 & 1.0& 2.0 & 10.7 & 19.0\\
        & GN16 & \textcolor{blue}{\textbf{0.8221$\pm$0.2055}} & 0.7721$\pm$0.2523 & 0.7664$\pm$0.2724& 0.5 & 1.5 & 1.5& 1.7 & 11.5 & 19.2\\
        & IN & 0.7942$\pm$0.2284 & \textcolor{blue}{\textbf{0.7895$\pm$0.2355}} & 0.7600$\pm$0.2443 & 0.5 & 1.5 & 0.5 & 1.7 & 7.1 & 13.7\\
        & LN            & 0.7586$\pm$0.2741 & 0.7122$\pm$0.3068 & 0.6992$\pm$0.3068 & 1.0 & 0.5 & 1.0   & 1.6 & 7.4 & 14.3\\
\hline
\end{tabular}
\label{tab:aorta} 
\end{table*}

\begin{table*}
\centering
\caption{Mean$\pm$std DSCs of the LV segmentation with different normalization methods (highest DSC in bold and blue).}
\begin{tabular}{|c|c|ccc|ccc|ccc|}
\hline
Experiment & Normalization & \multicolumn{3}{c|}{Mean$\pm$std DSCs} & \multicolumn{3}{c|}{Optimal LR} & \multicolumn{3}{c|}{Time (Seconds/20 Iterations)}\\
           &               & S-BS  &M-BS  &L-BS                     & S-BS & M-BS & L-BS                         & S-BS & M-BS & L-BS   \\
\hline
LV-1 & None          & 0.9240$\pm$0.0678 & 0.8344$\pm$0.2036 & 0.5277$\pm$0.3078 & 0.1 & 0.1 & 0.5 & 0.5 & 2.6 & 4.8\\
     & BN            & 0.9240$\pm$0.0808 & 0.9054$\pm$0.0864 & 0.8868$\pm$0.1432 & 0.5 & 0.5 & 0.5 & 0.7 & 3.0 & 5.5\\
     & GN4           & 0.9229$\pm$0.0979 & 0.8684$\pm$0.1552 & 0.8113$\pm$0.2102 & 0.5 & 1.5 & 0.1 & 1.1 & 5.3 & 9.4\\
     & GN8           & 0.9233$\pm$0.0864 & 0.8876$\pm$0.1346 & 0.7582$\pm$0.2196 & 1.0 & 1.0 & 0.5 & 1.1 & 5.1 & 9.5\\
     & GN16 &0.9306$\pm$0.0560 & \textcolor{blue}{\textbf{0.9233$\pm$0.0672}} & \textcolor{blue}{\textbf{0.9006$\pm$0.1058}}&0.1 & 1.0 & 1.0& 1.0 & 5.1 & 9.1\\
     & IN &\textcolor{blue}{\textbf{0.9313$\pm$0.0657}} & 0.9099$\pm$0.1064 & 0.8982$\pm$0.1230& 0.5 & 1.0 & 1.0& 1.0 & 3.9 & 7.3\\
     & LN            & 0.8426$\pm$0.1775 & 0.8552$\pm$0.1719 & 0.8531$\pm$0.1592 & 1.0 & 1.0 & 0.5 & 1.0 & 4.0 & 7.5\\
\hline      
LV-2 & None          & 0.8874$\pm$0.1592 & 0.7287$\pm$0.2496 & 0.5219$\pm$0.2830 & 0.1 & 0.05 & 0.1& 0.5 & 2.6 & 4.8\\
     & BN            & 0.8864$\pm$0.1391 & \textcolor{blue}{\textbf{0.8567$\pm$0.1815}} & 0.7892$\pm$0.2325 & 1.0 & 0.5 & 0.5 & 0.7 & 3.0 & 5.5\\
     & GN4 & \textcolor{blue}{\textbf{0.8931$\pm$0.1352}} & 0.8050$\pm$0.2018 & 0.7279$\pm$0.2128 & 0.5 & 0.5 & 1.0   & 1.1 & 5.3 & 9.4\\
     & GN8           & 0.8844$\pm$0.1161 & 0.8430$\pm$0.1439 & 0.7815$\pm$0.2229 & 1.0 & 0.1 & 0.1 & 1.1 & 5.1 & 9.5\\
     & GN16          & 0.8915$\pm$0.1288 & 0.8479$\pm$0.1736 & \textcolor{blue}{\textbf{0.8188$\pm$0.1608}}&1.0 & 0.5 & 1.0& 1.0 & 5.1 & 9.1\\
     & IN            & 0.8894$\pm$0.1459 & 0.8013$\pm$0.2428 & 0.7973$\pm$0.2138 & 0.5 & 0.5 & 1.0 & 1.0 & 3.9 & 7.3\\
     & LN            & 0.7806$\pm$0.2038 & 0.8389$\pm$0.1925 & 0.7059$\pm$0.1642 & 0.5 & 0.5 & 1.0 & 1.0 & 4.0 & 7.5\\
\hline      
LV-3 & None          & 0.8081$\pm$0.2345 & 0.6956$\pm$0.2828 & 0.6526$\pm$0.2932 & 0.1 & 0.05 & 0.5& 0.5 & 2.6 & 4.8\\
     & BN & \textcolor{blue}{\textbf{0.8479$\pm$0.1643}} & 0.7899$\pm$0.2186 & 0.7677$\pm$0.2151& 1.0 & 1.0 & 0.5& 0.7 & 3.0 & 5.5\\
     & GN4           & 0.8123$\pm$0.2355 & 0.7288$\pm$0.2526 & 0.7034$\pm$0.2515 & 0.5 & 1.5 & 0.1 & 1.1 & 5.3 & 9.4\\
     & GN8           & 0.8116$\pm$0.2297 & 0.7620$\pm$0.2508 & 0.7756$\pm$0.2422 & 0.1 & 1.5 & 1.0 & 1.1 & 5.1 & 9.5\\
     & GN16 & 0.8447$\pm$0.1882 & \textcolor{blue}{\textbf{0.8255$\pm$0.1982}} & \textcolor{blue}{\textbf{0.8013$\pm$0.2097}}&0.5 & 0.5 & 0.5& 1.0 & 5.1 & 9.1\\
     & IN            & 0.8401$\pm$0.1856 & 0.8044$\pm$0.2107 & 0.7660$\pm$0.2404 & 1.0 & 1.5 & 1.5 & 1.0 & 3.9 & 7.3\\
     & LN            & 0.7979$\pm$0.2437 & 0.7674$\pm$0.2555 & 0.6973$\pm$0.2699 & 0.1 & 0.1 & 1.0 & 1.0 & 4.0 & 7.5\\
\hline
\end{tabular}
\label{tab:LV} 
\end{table*}

\section{RESULTS}
\label{sec:result}
As stated in \Secref{sec:BN}, there are two ways of applying BN during the inference. Both of them are validated in \Secref{sec:inference}. The optimal results are selected to represent the BN performance and are used for later comparisons between different normalization methods in \Secref{sec:normmethods}. It is known that training the same model in multiple times would result in slightly different results \cite{sze2017efficient}. In this paper, the same phenomenon exists and the corresponding validations are in \Secref{sec:multiple}. Segmentation examples are illustrated in \Secref{sec:patient}. In the following paragraphs, RV-1 refers to the first cross validation (using the first group as the testing while using the second and third group as the training) for the RV, this name also applies to RV-2, RV-3, Aorta-1, Aorta-2, Aorta-3, LV-1, LV-2, LV-3. BS refers to batch size. S/M/L refers to the small/medium/large batch size, indicating batch size of (1, 16, 32), (1, 8, 16), (1, 16, 32) for the RV, aorta, LV respectively. LR refers to the learning rate. GN4, GN8, GN16 refers to the group normalization with group number of 4, 8, 16.

\subsection{Using BN during Inference}
\label{sec:inference}
TestI (using the mean and variance of the testing feature map to normalize the testing feature map) and TrainI (using the moving average mean and variance of the training feature maps to normalize the testing feature map) are validated on the three datasets with small, medium and large batch size. The mean$\pm$std DSCs are shown in \Tacref{tab:inference}. It is evident that TestI outperformed TrainI in most experiments, except those for the CT images (aorta) with medium and large batch sizes. However, this conclusion needs more systematic experiments to validate its generalization. In this paper, we use the optimal DSC achieved to represent the BN performance.

\subsection{Comparison between Normalization Methods}
\label{sec:normmethods}
To compare different normalization methods, U-Nets are trained with three datasets (RV, Aorta, LV), three cross validations, seven normalization methods (None, BN, GN4, GN8, GN16, IN, LN) and three batch sizes (small, medium large). The mean$\pm$std DSCs achieved are shown in \Tacref{tab:RV}, \ref{tab:aorta}, and \ref{tab:LV} respectively. It can be seen that for most experiments, GN16 or IN achieves the highest accuracy. For most exceptions, GN16 or IN could achieve similar accuracy to the highest value. As the feature root - C in this paper is 16, GN16 is similar to IN which divides the feature map into very small groups. It could be concluded that detailed subdivision of the feature map during normalization potentially leads to higher accuracy.

Adding normalization increases the running time. In general, BN is faster than IN, LN and GN. There is no rule regarding the LR. It is worth noting that training with small batch size outperformed that with large batch size. In the following paragraphs, we select experiments with small batch size for showing the convergence and patient errors.

\begin{figure*}[thpb]
\centering
\framebox{\includegraphics[scale=0.39]{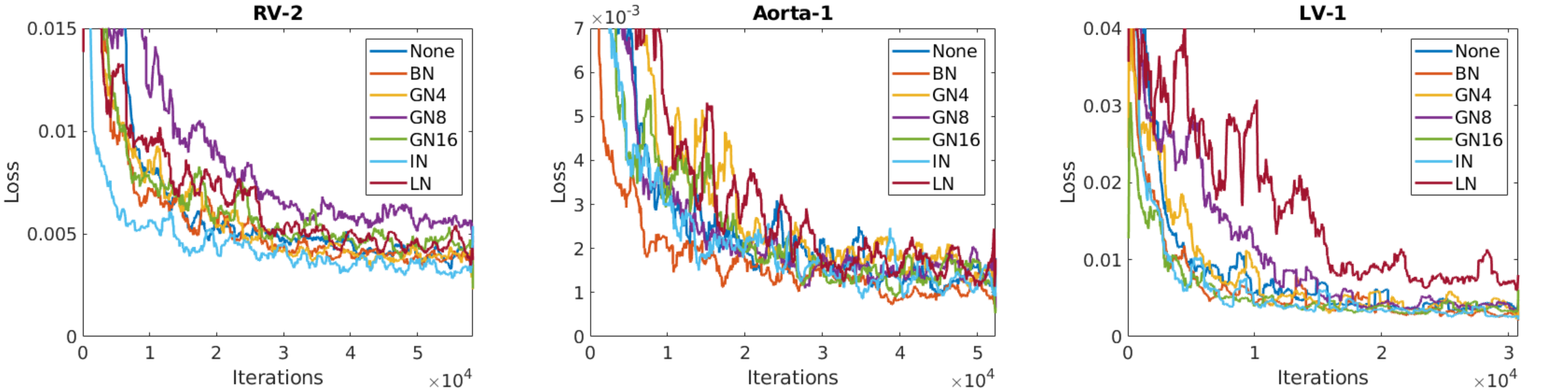}}
\caption{The training loss of U-Net with no normalization, BN, GN4, GN8, GN16, IN, and LN for RV-2 (left), aorta-1(middle), and LV-1(right) segmentation, the losses were recorded every 20 iterations, smoothed by a moving average window of 31, and truncated for clear plot.}
\label{fig:loss}
\end{figure*}

Three experiments - RV-2, Aorta-1, LV-1 are selected randomly to show the loss convergence during the training in \Ficref{fig:loss}. Unlike the report in \cite{ioffe2015batch} where the DCNN was trained 14 times faster, no obvious improvements on the convergence speed is observed. The optimal normalization methods for RV-2, Aorta-1 and LV-1 are GN16, GN16, and IN respectively. Obvious lower loss is achieved by GN16 for the RV-2 test while this phenomenon is not obvious for the Aorta-1 and LV-1 test. We think the validations are not enough to make a general conclusion.

\begin{figure}[thpb]
\centering
\framebox{\includegraphics[scale=0.62]{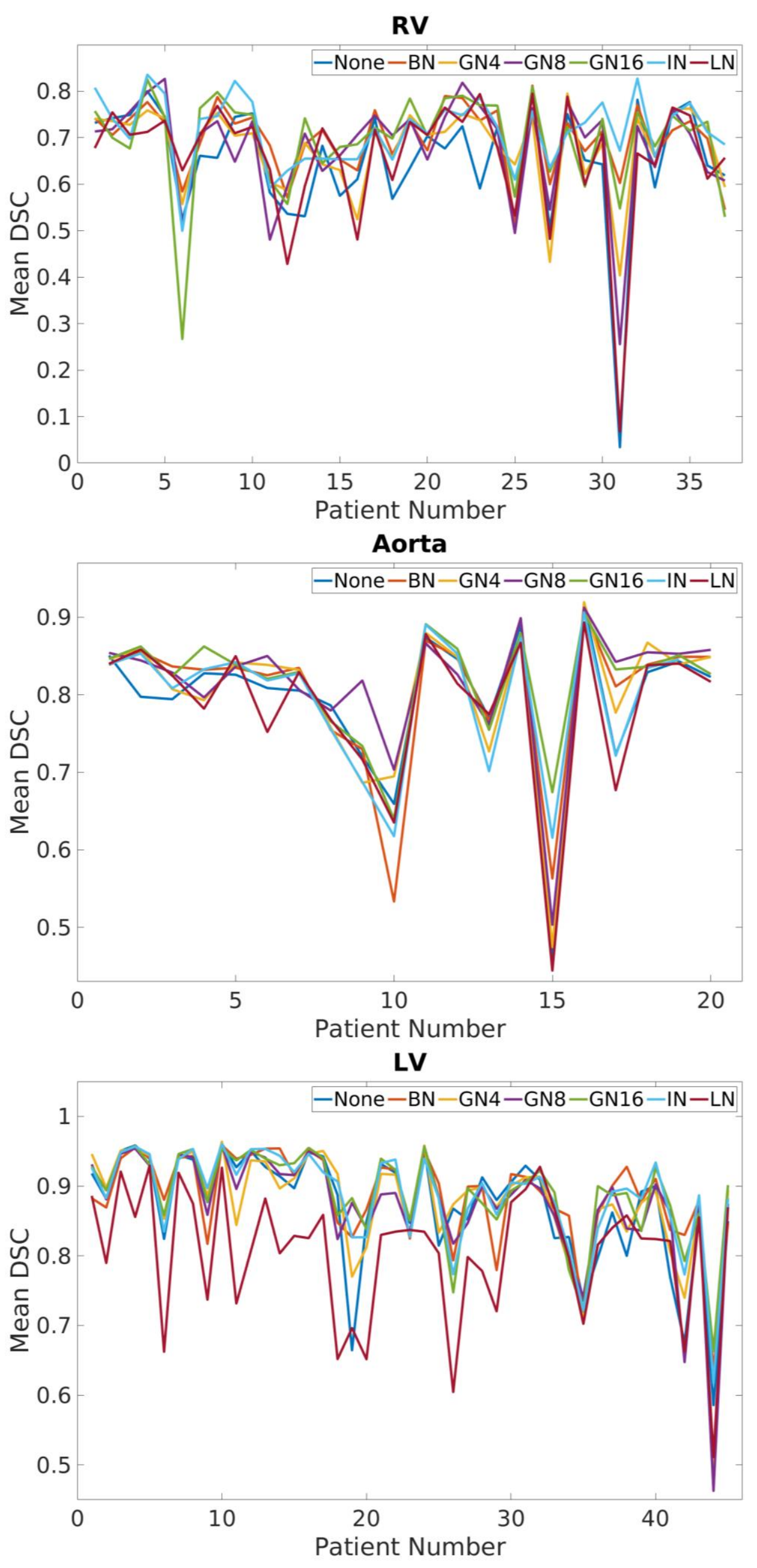}}
\caption{mean DSC for each patient for the RV (top), Aorta (middle), LV (bottom) segmented by U-Net with None, BN, GN4, GN8, GN16, IN, LN normalization methods.}
\label{fig:patient}
\end{figure}

We further show the mean DSC for each patient in \Ficref{fig:patient}. Due to the complex parameter setting when acquiring CT or MR images, the image intensity distributions are always different between patients. Hence, internal covariate shift is produced and some patients are with low segmentation accuracy. In \Ficref{fig:patient}, the main accuracy improvements achieved by normalization appear at those patients with low initial accuracy, i.e. patient 31 for RV, patient 15 for Aorta, patient 44 for LV. This proves that the accuracy improvement with normalization methods come from its improved generalization ability.

\begin{table}
\centering
\caption{Mean and std of the mean DSC when training the same model in six times.}
\begin{tabular}{|c|c|c|c|}
\hline
Normalization & Experiment & Mean DSC & Std DSC (6 runs) \\
\hline
None          & LV-1       & 0.8916   & 0.0248 \\
BN            & LV-3       & 0.8274   & 0.0129 \\
GN4           & RV-3       & 0.6436   & 0.0222 \\
GN8           & Aorta-2    & 0.7924   & 0.0097 \\
GN16          & Aorta-3    & 0.8091   & 0.0148 \\
IN            & RV-1       & 0.7035   & 0.0226 \\
LN            & LV-2       & 0.8189   & 0.0562 \\    
\hline
\end{tabular}
\label{tab:multiple} 
\end{table}

\subsection{Multiple Runs}
\label{sec:multiple}
It is known that training the same model in multiple times indicate different results - 2\% variance as stated in \cite{sze2017efficient}. In biomedical semantic segmentation, this exists as well. We select randomly one experiment for each normalization method and train it additionally five times. The mean and std of the mean DSCs of different trainings are shown in \Tacref{tab:multiple}. We can see that the $std$ between multiple runs is very large, sometimes can be even larger than the accuracy improvement. In this paper, all the results shown above are trained only once, this is fair for each method. However, running the experiments in multiple times may indicate a different results.

\subsection{Segmentation Results}
\label{sec:patient}
\begin{figure}[thpb]
\centering
\framebox{\includegraphics[scale=0.2]{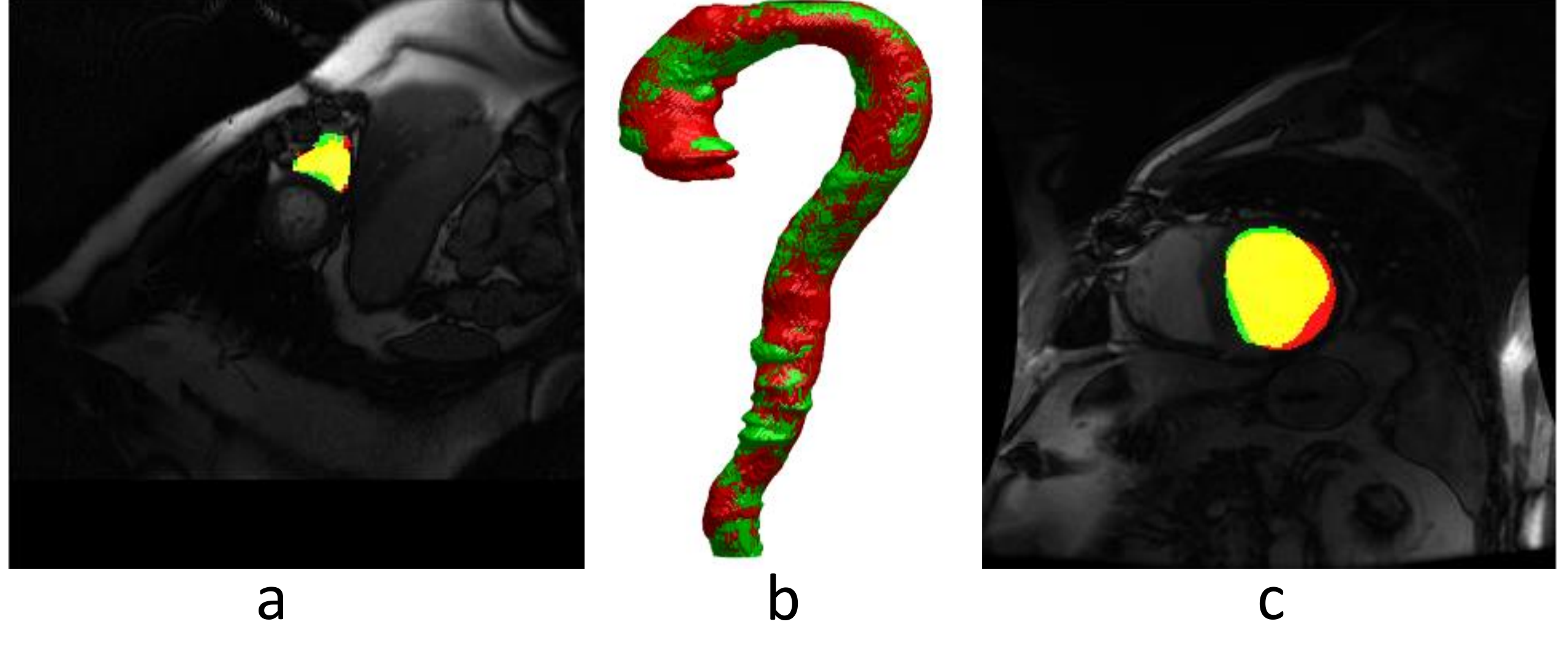}}
\caption{Segmentation examples of the RV (a), aorta (b) and LV (c). red - the ground truth, green - the segmentation results, yellow - the overlap between the ground truth and the segmentation results.}
\label{fig:seg}
\end{figure}

The 3D aortic shape reconstructed from the aortic segmentation is shown in \Ficref{fig:seg}b, which could be registered to navigate the Magellan (Hansen Medical, CA, USA) robotic system. As the RV and LV are MR images with 10mm slice gap, 3D reconstruction could not be extracted. The 2D RV and LV segmentation results are shown in \Ficref{fig:seg}a and \ref{fig:seg}c respectively, which could be used to instantiate 3D shapes and hence to navigate cardiac robotic interventions.

Two workstations with an Intel® Xeon(R) CPU E5-1650 v4 @ 3.60GHz × 12 and an Intel® Xeon(R) CPU E5-1620 v4 @ 3.50GHz × 8 were used for the training and testing. The GPUs used are Titan Xp with 12G memory and 1080 Ti with 11G memory. The time in \Tacref{tab:RV}, \ref{tab:aorta}, \ref{tab:LV} is recorded on the same computer and GPU for fairness.

\section{DISCUSSION}
\label{sec:discuss}

Most DCNNs for semantic segmentation applied BN as the normalization method. For biomedical semantic segmentation which is usually trained from scratch, it is possible to substitute the BN with other normalization methods for better performance. In this paper, we proved that detailed subdivision of the feature map, i.e. GN with a large group number or IN, facilitates the generalization of the trained model and hence improves the performance. Our experiments also indicate other conclusions: 1) small batch size out-performed large batch size; 2) TestI out-performed TrainI when applying BN during inference. However, we do not think our experiments are sufficient to fully prove these two conclusions. Hence, we would leave it open. Along with the publishing of this paper, new normalization methods are being proposed, i.e. \cite{nam2018batch} and \cite{wang2018kalman}. We are looking forward to future and further discussing. 

Although the focus of this paper is a fundamental problem in training U-Net for biomedical semantic segmentation - normalization, this paper connects and contributes to surgical robotic vision. The three segmented anatomies - RV, aorta, LV could be used for cardiac robotic navigation \cite{zhou2016path} and surgical robotic path planning, based on previous work of 3D shape instantiation \cite{zhou2018real}\cite{toth2015adaption}. 

\section{Conclusion}
\label{sec:conclusion}
In conclusion, this paper explores the biomedical semantic segmentation in surgical robotic vision and focuses on the normalization in training U-Nets. Four most popular normalization methods - BN, IN, LN and GN are reviewed and compared in details. Detailed subdivision of the feature map, i.e. GN with a large group number or IN, improves the accuracy of training U-Net for biomedical semantic segmentation. This accuracy improvement is mainly from improved generalization ability of the trained model. This paper could help with indicating the future direction on proposing new normalization methods.

\section*{Acknowledgment}
This work is supported by the EPSRC project grant EP/L020688/1. We gratefully acknowledge the support of NVIDIA Corporation with the donation of the Titan Xp GPU used for this research. Thank Qing-Biao Li for the data processing. Thank Jian-Qing Zheng for some figures.

\bibliographystyle{IEEEtran}
\bibliography{ICRA2019}
\end{document}